\documentclass[10pt,twocolumn,letterpaper]{article}

\usepackage{iccv}
\usepackage{times}
\usepackage{epsfig}
\usepackage{graphicx}
\usepackage{amsmath}
\usepackage{amssymb}
\usepackage{booktabs}
\usepackage[nocompress]{cite}

\usepackage[pagebackref=true,breaklinks=true,letterpaper=true,colorlinks,bookmarks=false]{hyperref}

\iccvfinalcopy 

\def\iccvPaperID{12511} 

\ificcvfinal\fi

\begin{document}

\title{Multi-Similarity Contrastive Learning}

\author{%
  Emily Mu \\
  Massachusetts Institute of Technology\\
  \texttt{emilymu@mit.edu}\\
  \and
  John Guttag \\
  Massachusetts Institute of Technology\\
  \texttt{guttag@csail.mit.edu}\\
  \and
  Maggie Makar \\
  University of Michigan \\
  \texttt{mmakar@umich.edu}\\
}

\maketitle
\ificcvfinal\thispagestyle{empty}\fi

\begin{abstract}
Given a similarity metric, contrastive methods learn a representation in which examples that are similar are pushed together and examples that are dissimilar are pulled apart. Contrastive learning techniques have been utilized extensively to learn representations for tasks ranging from image classification to caption generation. However, existing contrastive learning approaches can fail to generalize because they do not take into account the possibility of different similarity relations. In this paper, we propose a novel multi-similarity contrastive loss (MSCon), that learns generalizable embeddings by jointly utilizing supervision from multiple metrics of similarity. Our method automatically learns contrastive similarity weightings based on the uncertainty in the corresponding similarity, down-weighting uncertain tasks and leading to better out-of-domain generalization to new tasks. We show empirically that networks trained with MSCon outperform state-of-the-art baselines on in-domain and out-of-domain settings.
\end{abstract}

\section{Introduction}
\label{sec:intro}

Contrastive methods learn embeddings by pushing similar examples together and pulling dissimilar examples apart. Embeddings trained using contrastive learning have been shown to achieve state-of-the-art performance on a variety of computer vision tasks \cite{radford2021learning, yuan2021florence, khosla2020supervised}. In contrastive learning, representations are trained to discriminate pairs of similar images (positive examples) from a set of dissimilar images (negative examples). Supervised contrastive learning approaches consider all instances with the same label to be positive examples and all examples with different labels to be negative examples \cite{khosla2020supervised}.

Existing contrastive learning methods can fail to generalize because the learned embeddings are too simplistic, reflecting limited similarities between different examples. This limitation exists because current contrastive learning methods only consider a single way of defining similarity between examples. In settings where multiple notions of similarity are available, relying on only one notion of similarity represents a missed opportunity to learn more general representations \cite{khosla2020supervised, islam2021broad}. A challenge of generalizing to multiple notions is that training using multiple tasks adds complexity when tasks have different levels of uncertainty. Incorporating noisy similarity measures can lead to worse generalization performance. In multi-task and meta learning, it has been demonstrated that assigning different weights based upon relative task uncertainty can help models focus on tasks with low uncertainty, potentially leading to better classification accuracy and generalization towards new tasks and datasets \cite{kendall2018multi, mao2022metaweighting}. In this work, we propose multi-similarity contrastive loss (MSCon), a novel loss function that utilizes supervision from multiple similarity metrics and learns to down-weight more uncertain similarities. 

\begin{figure}[ht]
\begin{center}
\centerline{\includegraphics[width=\columnwidth]{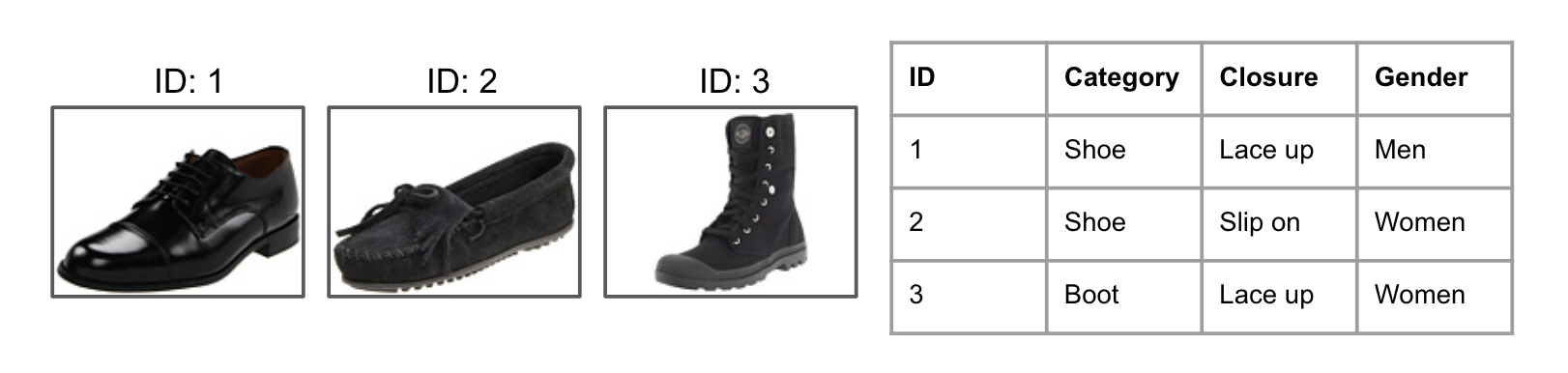}}
\caption{\textbf{Shoe Example.} An example illustrating multiple disjoint similarity relationships between three images of shoes.}
\label{fig:tab}
\end{center}
\end{figure}

Throughout, we will use shoe classification as a motivating example. Each of the shoes in Figure~\ref{fig:tab} is associated with distinct category, closure, and gender attributes. For example, images 1 and 2 are similar in category but are dissimilar in closure and gender, while images 2 and 3 are similar in gender but dissimilar in category and closure. We refer to such a dataset as a multi-similarity dataset. Other examples of multi-similarity datasets include multiple disease labels associated with chest radiographs \cite{irvin2019chexpert, johnson2020mimic} and relational tables associated with website text \cite{chen2000mining, bhagavatula2015tabel}. For convenience, we will refer to the similarity function induced by the labels of a task as the similarity metric of that task.

Suppose we are training a model using all three tasks: category, closure, and gender. Closure might be a task with low noise and low uncertainty, while gender might be a task with high noise and higher uncertainty. We find that our approach learns a higher weight for closure than for gender, ensuring that the model focuses more on closure during training.

\begin{figure}[t!]
\begin{center}
\centerline{\includegraphics[width=\columnwidth]{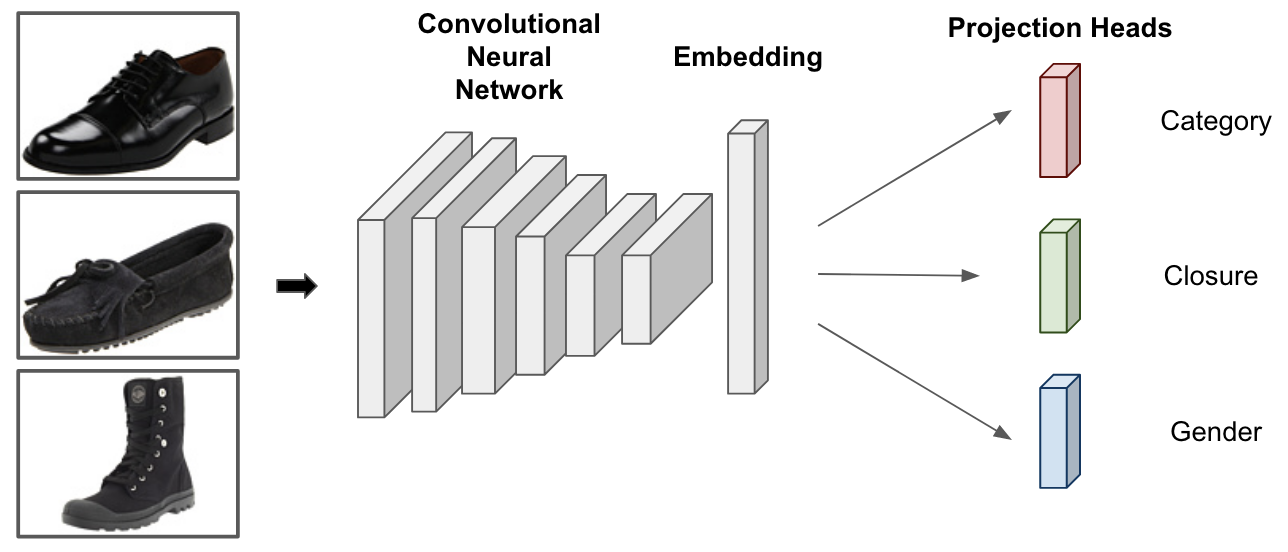}}
\caption{\textbf{Multi-Similarity Contrastive Network.} Multiple projection heads are trained to learn from multiple metrics of similarity. The base encoding network and the projection heads are trained together. The projection heads are discarded and only the encoding network is kept for downstream tasks. During training, our network is able to learn weightings for each similarity metric based on the task uncertainty.}
\label{fig:ccn}
\end{center}
\end{figure}

Our framework is shown in Figure~\ref{fig:ccn}. MSCon uses multiple projection heads to learn embeddings based on different metrics of similarity. In this way, we are able to represent examples that are positive examples in one projected subspace and negative examples in a different projected subspace. Additionally, we model similarity-dependent uncertainty by first constructing a pseudo-likelihood function. Since our contrastive loss uses a non-parametric approach to learn the similarities between two inputs, we use the pseudo-likelihood function to approximate label uncertainty in the learned similarity spaces. We then learn a weighting parameter for each similarity metric that maximizes this pseudo-likelihood. 

In extensive experiments, we show that our weighting scheme allows models to learn to down-weight more uncertain similarity metrics, which leads to better generalization of the learned representation to novel tasks. We also show that embeddings trained with our multi-similarity contrastive loss outperform embeddings trained with traditional self-supervised and supervised contrastive losses on two multi-similarity datasets. Finally, we show that embeddings trained with MSCon generalize better to out-of-domain tasks than do embeddings trained with multi-task cross-entropy.

Our main contributions are:

\vspace{-\topsep}

\begin{enumerate}
    \setlength{\parskip}{-0pt}
    \setlength{\itemsep}{0pt plus 1pt}
    \item We propose a novel multi-similarity contrastive learning method for utilizing supervision based on multiple metrics of similarity.
    \item We propose a weighting scheme to learn robust embeddings in the presence of possibly uncertain similarities induced by noisy tasks. Our weighting scheme learns to down-weight uninformative or uncertain tasks leading to better out-of-distribution generalization.
    \item We empirically demonstrate that a network trained with our multi-similarity contrastive loss performs well for both in-domain and out-of-domain tasks and generalizes better than multi-task cross-entropy methods to out-of-domain tasks.
\end{enumerate}

\section{Related Work}\label{sec:related}

\subsection{Contrastive Representation Learning}

Our work draws from existing literature in contrastive representation learning. Many of the current state-of-the-art vision and language models are trained using contrastive losses \cite{radford2021learning, yuan2021florence, chen2020simple, khosla2020supervised, he2020momentum}. Self-supervised contrastive learning methods, such as MoCo and SimCLR, maximize agreement between two different augmentations or views of the same image \cite{he2020momentum, chen2020simple}. Recently, vision-language contrastive learning has allowed dual-encoder models to pretrain with hundreds of millions of image-text pairs \cite{jia2021scaling, radford2021learning}. The resulting learned embeddings achieve state-of-the-art performance on many vision and language benchmarks \cite{yuan2021florence, wang2022git, li2022grounded}. Supervised contrastive learning, SupCon, allows contrastive learning to take advantage of existing labels \cite{khosla2020supervised}. Contrastive learning has also been adapted to learn from both labels and text \cite{yang2022unified} and from hierarchies of labels \cite{zhang2022use}. The method most similar to ours is conditional similarity networks \cite{veit2017conditional}. In conditional similarity networks, masks are learned or assigned to different embedding dimensions with respect to different metrics of similarity. These masks are learned jointly with the convolutional neural network parameters during training time. Conditional similarity networks differ from our work in two major ways. First, unlike our work, conditional similarity networks uses triplet loss, a specialized version of contrastive loss. At training time, it requires triplets based on each similarity metric. Second, we automatically learn separate projection spaces and weights for each metric of similarity, whereas they learn a linear transformation from the embedding space for each similarity and do not consider weighting metrics. As we show in Section~\ref{ssec:classification}, our multiple similarity contrastive networks consistently outperforms conditional similarity networks.

\subsection{Multi-Task Learning}

Multi-task learning aims to simultaneously learn multiple related tasks and often outperforms learning each task alone \cite{kendall2018multi, bhattacharjee2022mult, mao2020multitask}. However, if tasks are weighted improperly during training, the performance on some tasks suffer. Various learned task weighting methods have been proposed for multi-task learning in the vision and language domains \cite{mao2022metaweighting, kendall2018multi, chen2018gradnorm, sener2018multi, liu2017adversarial, mao2020multitask, mao2021banditmtl}. These methods learn task weightings based on different task characteristics in order to improve the generalization performance towards novel tasks \cite{mao2022metaweighting}. This is done by regularizing the task variance using gradient descent \cite{chen2018gradnorm, mao2021banditmtl} or by using adversarial training to divide models into task-specific and generalizable parameters \cite{liu2017adversarial}. Overwhelmingly, these methods are built for multiple tasks trained with likelihood-based losses, such as regression and classification. One of the most popular of these methods models task uncertainty to determine task-specific weighting and automatically learns weights to balance this uncertainty \cite{kendall2018multi}. In our work, we adapt automatically learned task weighting to our multi-similarity contrastive loss by predicting similarity uncertainty. This is not straightforward since the contrastive loss is trained in a pairwise fashion and there is a lack of absolute labels in the learned output (a set of embedding vectors) \cite{ardeshir2022uncertainty}.

\subsection{Uncertainty in Contrastive Learning}

Adapting uncertainty estimation techniques to contrastive learning remains an active area of research. This is because contrastive learning learns abstract embedding vectors rather than absolute labels and because of the pairwise training of contrastive models. Given access to training data and labels, previous work has proposed estimating the density and consistency of the hypersphere embedding space distribution as metrics to estimate uncertainty \cite{ardeshir2022uncertainty, oh2018modeling}. The density of the embedding space at a point captures the amount of data the model has observed during training, and can serve as a proxy for epistemic, or model, uncertainty \cite{oh2018modeling}. The consistency of the embedding space at a point uses k-nearest-neighbors to measure the extent to which the training data mapped closest to that point have consistent labels, and can serve as a proxy for aleatoric, or data-dependent, uncertainty \cite{ardeshir2022uncertainty}. Other recent work proposes learned temperature as a metric of heteroscedastic, or input-dependent, uncertainty to identify out-of-distribution data for labeled datasets \cite{zhang2021temperature}. To our knowledge, we are the first work to model similarity-dependent uncertainty, or the relative confidence between different training tasks, in the contrastive setting.

\section{Method}\label{sec:method}

\subsection{Multi-Similarity Setup} 

We assume that during training time, we have access to dataset: $\mathcal{D} = \{x_i, \textbf{Y}_i\}_i^M$, where $x$ is an image and the $\textbf{Y}_i = \{y_i^1... y_i^C\}$ are distinct categorical attributes  associated with the image. We aim to learn an embedding function $f: x \rightarrow \mathbb{R}^{d}$ that maps $x$ to an embedding space. We define $h_i = f(x_i)$ to be the embedding of $x_i$. 

In the typical contrastive training setup, training proceeds by selecting a batch of $N$ randomly sampled data $\{x_i\}_{i=1...N}$. We randomly sample two distinct label preserving augmentations (e.g., from rotations, crops, flips) for each $x_i$, ($\tilde{x}_{2i}$ and $\tilde{x}_{2i-1}$), to construct $2N$ augmented samples, $\{\tilde{x}_{j}\}_{j=1...2N}$. Let $A(i) = \{1,... 2N\} \backslash i$ be the set of all samples and augmentations not including $i$. We define $g$ to be a projection head that maps the embedding to the similarity space represented as the surface of the unit sphere $\mathbb{S}^{d} = \{v \in \mathbb{R}^d: ||v||_2=1\}$. Finally, we define $v_i = g(h_i)$ as the mapping of $h_i$ to the projection space. 

Supervised contrastive learning uses labels to implicitly define the positive sets of examples. Specifically, supervised contrastive learning encourages samples with the same label to have similar embeddings and samples with a different label to have different embeddings. We follow the literature in referring to samples with the same label as an image $i$ as the positive samples, and samples with a different label than that of $i$'s as the negative samples. 

Supervised contrastive learning (SupCon) \cite{khosla2020supervised} proceeds by minimizing the loss:

\vspace{-4mm}

\begin{equation}
\begin{footnotesize}
L^{supcon} = \sum_{i=I} \frac{-1}{|P(i)|} \sum_{p \in P(i)} \log \frac{\text{exp}(\frac{v_i^T v_p}{\tau})}{\sum_{a \in A(i)} \text{exp}(\frac{v_i^T v_a}{\tau})}, 
\label{eq:supcon}
\end{footnotesize}
\end{equation}
where $|S|$ denotes the cardinality of the set $S$,  $P(i)$ denotes the positive set with all other samples with the same label as $x_i$, i.e., $P(i) = \{j \in A(i): y_j = y_i\}$, $I$ denotes the set of all samples in a particular batch, and $\tau \in \{0, \infty\}$ is a temperature hyperparameter.

In contrast to SupCon, our multi-similarity contrastive (MSCon) approach proceeds by jointly training an embedding space using multiple notions of similarity. We do so by training the embedding with multiple projection heads $g^c$ that map the embedding to $C$ projection spaces, where each space distinguishes the image based on a different similarity metric. We define $v^c_i = g^c(h_i)$ to be the mapping of $h_i$ to the projection space by projection head $g^c$. Because each projection space is already normalized, we assume that the each similarity loss is similarly scaled. We define the multi-similarity contrastive loss to be a summation of the supervised contrastive loss over all conditions $L^{mscon} = \sum_{c \in C} \sum_{i=I} L^{mscon}_{c,i}
$ where each conditional $L^{mscon}_{c,i}$ is defined as in equation~\ref{eq:mscon}. Specifically,

\vspace{-4mm}

\begin{equation}
\begin{footnotesize}
L^{mscon}_{c,i} = \frac{-1}{|P^c(i)|} \sum_{p \in P^c(i)} \log \frac{\text{exp}(\frac{v_i^{cT} v^c_p}{\tau})}{\sum_{a \in A(i)} \text{exp}(\frac{v_i^{cT} v^c_a}{\tau})},
\label{eq:mscon}
\end{footnotesize}
\end{equation}

where $P^c(i)$ is defined as the positive set under similarity $c$ such that for all $j \in P^c(i)$, $y_j^c = y_i^c$. 

\subsection{Contrastive Task Weighting}

In the above formulation of our multi-similarity contrastive loss function, each similarity is weighted equally. However, previous work in multi-task learning for both vision and language have demonstrated that model performance can deteriorate when one or more of the tasks is noisy or uncertain. One way to tackle this is to learn task weights based on the uncertainty of each task. However, model performance can be sensitive to weight selection \cite{gong2019comparison, mao2022metaweighting, kendall2018multi, chen2018gradnorm, mao2021banditmtl}, and manually searching for optimal weightings is expensive in both computation and time. Previous work has suggested using irreducible uncertainty of task predictions in a weighting scheme. For example, tasks where predictions are more uncertain are weighted lower because they are less informative\cite{kendall2018multi}.

Such notions of uncertainty are typically predicated on an assumed parametric likelihood of a label given inputs. However, this work is not easily adapted to multi-similarity contrastive learning because 1) contrastive training does not directly predict downstream task performance and 2) the confidence in different similarity metrics has never been considered in this setting. In contrastive learning, the estimate of interest is a similarity metric between different examples rather than a predicted label, which means that downstream task performance is not directly predicted by training results. Furthermore, previous work in contrastive learning has only focused on modeling data-dependent uncertainty, or how similar a sample is to negative examples within the same similarity metric. To our knowledge, we are the first to utilize uncertainty in the training tasks and their corresponding similarity metrics as a basis for constructing a weighting scheme for multi-similarity contrastive losses.

We do this in two ways: 1) we construct a pseudo-likelihood function approximating task performance and 2) we introduce a similarity dependent temperature parameter to model relative confidence between different similarity metrics. We present an extension to the contrastive learning paradigm that enables estimation of the uncertainty in similarity metrics. In addition to providing useful information about the informativeness of each similarity metric, our estimate of uncertainty enables us to weight the different notions of similarity such that noisy notions of similarity are weighted lower than more reliable notions. 

Our approach proceeds by constructing a pseudo-likelihood function which approximates task performance. We show in the supplement that maximizing our pseudo-likelihood also maximizes our MSCon objective function. This pseudo-likelihood endows the approach with a well-defined notion of uncertainty that can then be used to weight the different similarities.  

Let $v_i^c$ be the model projection head output for similarity $c$ for input $x_i$. Let $\textbf{Y}^c$ be the $c$th column in $\textbf{Y}$. We define $P^c_y = \{x_j \in \mathcal{D} : \textbf{Y}^c_j = y\}$ to be the positive set for label $y$ under similarity metric $c$. We define the classification probability $p(y|v_i^c, D, \tau)$ as the average distance of the representation $v_i^c$ from all representations for inputs conditioned on the similarity metric. Instead of directly optimizing equation~\ref{eq:supcon}, we can maximize the following pseudo-likelihood: 

\vspace{-4mm}

\begin{equation}
\begin{footnotesize}
p(y|v_i^c, D, \tau) \propto \frac{1}{|P^c_y|} \sum_{p \in P^c_y} \text{exp}(\frac{v_i^{cT} v_p^c}{\tau}).
\end{footnotesize}
\label{eq:class-prob}
\end{equation}

Note that optimizing~\ref{eq:class-prob} is equivalent to optimizing~\ref{eq:supcon} by applying Jensen's inequality (as shown in the supplement). By virtue of being a pseudo-likelihood, equation~\ref{eq:class-prob} provides us with a well-defined probability associated with downstream task performance that we can use to weight the different tasks. We will next outline how to construct this uncertainty from the pseudo-likelihood defined in equation~\ref{eq:class-prob}. 

We assume that $v^c$ is a sufficient statistic for $y^c$, meaning that $y^i$ is independent of all other variables conditional on $v^i$. Such an assumption is not unrealistic, it simply reflects the notion that $v^c$ is an accurate estimation for $y^c$. Under this assumption the pseudo-likelihood expressed in~\ref{eq:class-prob} factorizes as follows:

\vspace{-4mm}

\begin{equation}
\begin{footnotesize}
p(y^1, ... y^C|v_i^1, ... v_i^C, D, \tau) = p(y^1|v_i^1, D, \tau) ... p(y^C|v_i^C, D, \tau).
\end{footnotesize}
\label{eq:mult-fact}
\end{equation}

Previous work in contrastive learning modifies the temperature to learn from particularly difficult data examples \cite{zhang2021temperature, robinson2020contrastive}. Inspired by this, we adapt the contrastive likelihood to incorporate a similarity-dependent scaled version of the temperature. We introduce a parameter $\sigma_c^2$ for each similarity metric controlling the scaling of temperature and representing the similarity dependent uncertainty in Equation~\ref{eq:uncertainty}.

\vspace{-4mm}

\begin{equation}
\begin{footnotesize}
p(y|v_i^c, D, \tau, \sigma_c^2) \propto \frac{1}{|P^c_y|} \sum_{p \in P^c_y} \text{exp}(\frac{v_i^{cT} v_p^c}{\tau \sigma_c^2})
\end{footnotesize}
\label{eq:uncertainty}
\end{equation}

The negative log-likelihood for this contrastive likelihood can be expressed as Equation~\ref{eq:log-likelihood}.

\vspace{-4mm}

\begin{equation} \label{eq:log-likelihood}
\begin{aligned}
- \text{log } p(y|v_i^c, D, \tau, \sigma_c^2)
&\propto \frac{1}{\sigma_c^2} \sum_{i=I} L^{mscon}_{c,i} + 2\text{log}(\sigma_c)
\end{aligned}
\end{equation}

We provide a detailed derivation of this equation in the supplement. Extending this analysis to consider multiple similarity metrics, we can adapt the optimization objective to learn weightings for each similarity as in Equation~\ref{eq:mscon-w}.

\vspace{-4mm}

\begin{equation} \label{eq:mscon-w}
\begin{footnotesize}
\text{argmin}_{f, g_1, ... g_C, \sigma_1, ... \sigma_C} (\sum_{c \in C} (\frac{1}{\sigma_c^2} \sum_{i=I} L^{mscon}_{c,i} +  2\text{log}(\sigma_c)))
\end{footnotesize}
\end{equation}

During training, we learn the $\sigma_c$ weighting parameters through gradient descent.

\section{Experiments}\label{sec:exp}

In this section, we evaluate the performance of our approach: 1) under varying levels of uncertainty in similarity metrics induced by varying levels of task noise and 2) across in-domain and out-of-domain classification tasks. We show that our multi-similarity contrastive loss significantly outperforms existing self-supervised and single-task supervised contrastive networks and outperforms multi-task cross-entropy networks on novel tasks. We also demonstrate that our method is able to learn to down-weight more uncertain similarities, and that compared to using equal weights, our weighted multi-similarity contrastive loss is more robust to similarity metric uncertainty and generalizes better to novel tasks under increasing uncertainty.

\subsection{Datasets and Implementation} 

\paragraph{Datasets.} We use two datasets: Zappos50k \cite{yu2014fine, yu2017semantic} and MEDIC \cite{alam2022medic, alam2018crisismmd, alam2020deep, mouzannar2018damage, nguyen2017damage}. Sample images are provided in the supplement. 

\textbf{Zappos50k} consists of 50,000 $136 \times 102$ images of shoes. We focus our analysis on three tasks: the category of shoe (shoes, boots, sandals, or slippers), the suggested gender of the shoe (for women, men, girls, boys), and the closing mechanism of the shoe (buckle, pull on, slip on, hook and loop, or laced). We fine-tune the embedding space to predict the brand of the shoe for the out-of-domain experiment. We split the images into 70\% training, 10\% validation, and 20\% test sets and resize all images to $112 \times 112$.

\textbf{MEDIC} is the largest multi-task learning disaster-related dataset, extending the CRISIS multi-task image benchmark dataset \cite{alam2018crisismmd, alam2022medic}. MEDIC consists of $\approx71,000$ images of disasters collected from Twitter, Google, Bing, Flickr, and Instagram. The dataset includes four disaster-related tasks that are relevant for humanitarian aid: the disaster type (earthquake, fire, flood, hurricane, landslide, other disaster, and not a disaster), the informativeness of the image for humanitarian response (informative or not informative), categories relevant to humanitarian response (having affected, injured, or dead people, infrastructure and utility damage, rescue volunteering or donation effort, and not needing humanitarian response), and the severity of the damage of the event (severe damage, mild damage, and little to no damage). For the out-of-domain analysis, we hold out each task from training and then attempt to predict the hold-out task during evaluation. These tasks were generated from a crowd sourcing annotation platform and the images are split already into 69\% training, 9\% validation and 22\% test sets. All images were resized to $224 \times 224$.

\paragraph{Training Details.}

Consistent with previous work \cite{chen2020simple, khosla2020supervised}, images are augmented by applying various transformations to increase dataset diversity. We train using standard data augmentations, including random crops, flips, and color jitters.

An embedding network consisting of a shared encoder and multiple projection heads is then trained using MSCon with multiple similarity metrics defined by different tasks as shown in Figure~\ref{fig:ccn}. The resulting vectors are normalized to the unit hypersphere, which allows us to use an inner product to measure distances in the projection space. Zappos50k encoders use ResNet18 backbones with projection heads of size 32. MEDIC encoders use ResNet50 backbones with projection spaces of size 64 \cite{he2016deep}. All models are pretrained on ImageNet \cite{deng2009imagenet}. All networks are trained using a SGD with momentum optimizer for 200 epochs with a batch-size of 64 and a learning rate of 0.05, unless otherwise specified. We use a temperature of $\tau = 0.1$.

After training the multi-similarity contrastive network, we discard the projection heads and freeze the encoder network. We then evaluate the performance of the embedding network on downstream tasks by training a linear classifier on the embedding features. We train a linear classifier for 20 epochs and evaluate top-1 accuracy. Standard deviations are computed by bootstrapping the test set 1000 times. Additional implementation details can be found in the supplement. We will release code for implementing MSCon.

\paragraph{Models.} We compare the unweighted and weighted versions of our \textbf{Multi-Similarity Contrastive Network (MSCon)} with the following baselines:

\begin{itemize}

\setlength{\parskip}{0pt}
\vspace{-5pt}%
    \item \textbf{Cross-Entropy Networks (XEnt)} We train separate cross-entropy networks with each of the available tasks. We also train a multitask cross-entropy network with all available tasks. We train each network with a learning rate of 0.01. We select the best model using the validation accuracy. 
    \item \textbf{Conditional Similarity Network (CSN)} We train a conditional similarity network that learns the convolutional filters, embedding, and mask parameters together. 10,000 triplets are constructed from the similarities available in the training dataset. We follow the training procedure specified in \cite{veit2017conditional}. 
    \item \textbf{SimCLR and SupCon Networks} We train a self-supervised contrastive network for each dataset and individual supervised contrastive networks with each of the similarity metrics represented in the training dataset. We pretrain with a temperature of 0.1 for all contrastive networks which is the typical temperature used for SimCLR and SupCon \cite{chen2020simple, khosla2020supervised}. For evaluation, we fine-tune a classification layer on the frozen embedding space.
\end{itemize}

\subsection{Role of Weighting in Achieving Robustness to Task Uncertainty}

\begin{figure*}[ht!]
\vskip -0.1in
\begin{center}
\centerline{\includegraphics[width=2\columnwidth]{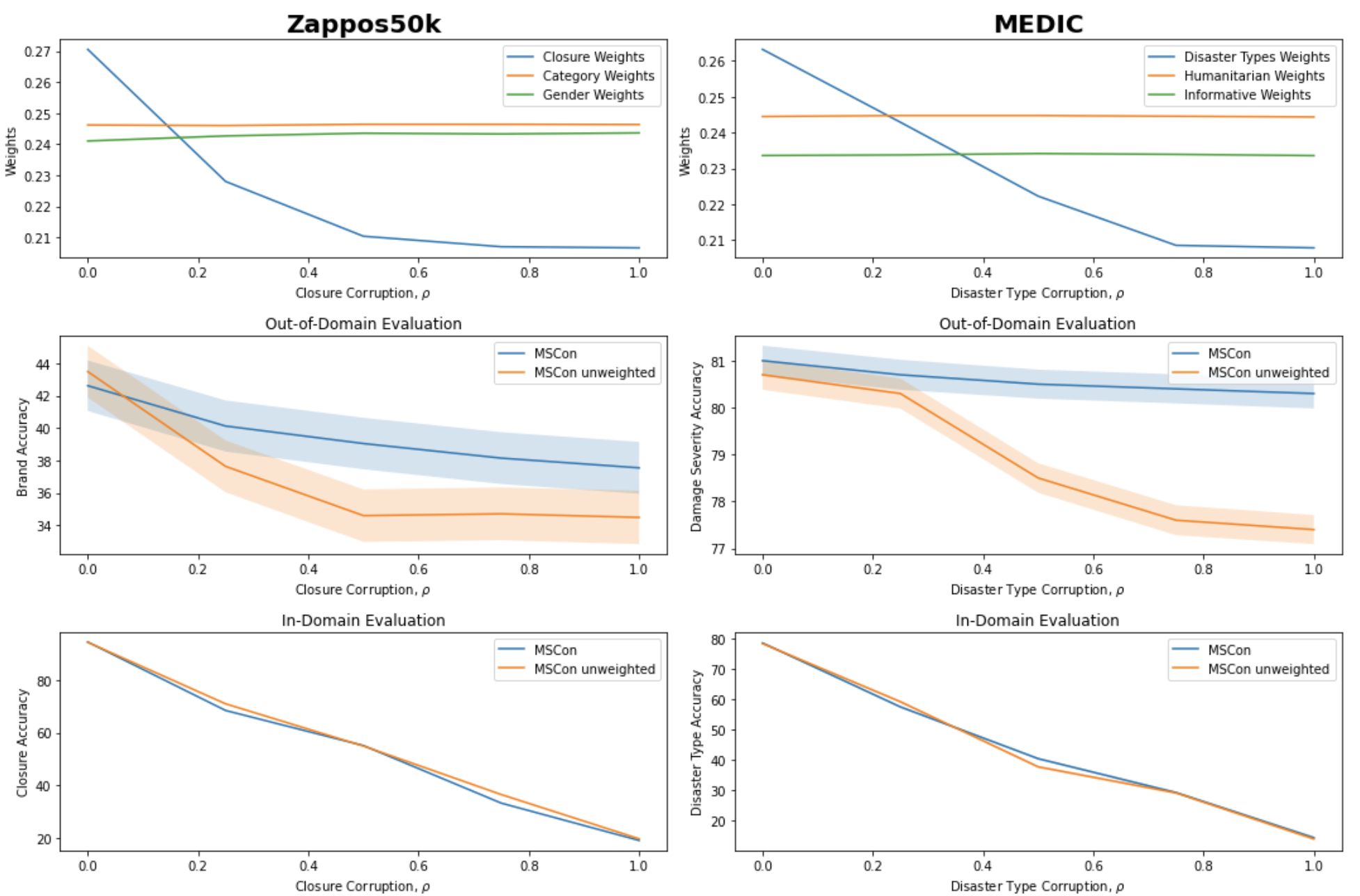}}
\caption{\textbf{Weighted MSCon generalizes better to unseen tasks under increasing similarity task uncertainty.} Unweighted and weighted versions of MSCon are trained on increasing task corruption. The x-axis on all plots represents the amount of task corruption $\rho$. The top row plots show that weighted MSCon learns to downweight the corrupted task. For both the Zappos50k and MEDIC datasets, weighted MSCon generalizes better to out-of-domain tasks than unweighted MSCon under increasing values of $\rho$, even as in-domain performance on the corrupted task degrades to random selection.}
\label{fig:corruption}
\end{center}
\end{figure*}

In this subsection, we evaluate the robustness of our learned embeddings to similarity uncertainty. Since the true level of task noise (similarity metric uncertainty) is unobserved, we use a semi-simulated approach, where we simulate uncertain similarities in both the Zappos50k and MEDIC datasets. 

For the Zappos50k dataset, we train the encoder using the category, closure, and gender similarity metrics. To introduce task uncertainty, we randomly corrupt the closure task by proportion $\rho$. We randomly sample $\rho$ of the closure labels and randomly reassign the label amongst all possible labels. Note that when $\rho = 1.0$, all labels are randomly sampled equally from the available closure labels. When $\rho = 0.0$, all labels are identical to the original dataset. For the MEDIC dataset, we train the encoder using the disaster types, humanitarian, and informative similarity metrics. We corrupt the disaster type task in order to introduce task uncertainty. As $\rho$ increases in Figure~\ref{fig:corruption}, we find that MSCon learns to down-weight the noisy task for both the Zappos50k and MEDIC datasets.

For the Zappos50k dataset, we evaluate the top-1 classification accuracy on an out-of-domain task, brand classification, and on an in-domain task, the corrupted closure classification. Similarly, for the MEDIC dataset, we evaluate the top-1 classification accuracy on an out-of-domain task, damage-severity classification, and on an in-domain task, the corrupted disaster-type classification. 

Figure~\ref{fig:corruption} shows the results from this analysis. The top panel shows how the weights change as we change task uncertainty on the x-axis. The middle and bottom panels shows how out-of-domain and in-domain evaluation accuracy changes as we change task uncertainty. As expected, as $\rho$ increases to $1$, the in-domain classification accuracy for both the equal-weighted and weighted MSCon learned embeddings decreases to random. 

However, the out-of-domain classification accuracy for the weighted MSCon learned embeddings is more robust to changes in $\rho$ than the unweighted MSCon learned embeddings. This is because the weighted version of MSCon automatically learns to down-weight uncertain or more uninformative tasks during encoder training.

\subsection{Classification Performance} \label{ssec:classification}

In this section, we evaluate in- and out-of-domain performance of various methods. We find that our multi-similarity contrastive network significantly outperforms all other contrastive methods on in-domain tasks and outperforms multi-task cross-entropy learning on out-of-domain tasks. We also show how performance changes with variation in hyperparameter selection. More qualitative analysis of the learned similarity subspaces (i.e., TSNE visualizations) can be found in the supplement.

\paragraph{In-domain Performance.} To evaluate the quality of the learned embedding spaces, we measure top-1 classification accuracy on all tasks for both the Zappos50k and MEDIC datasets. We report the average accuracy and the standard deviation for all tasks in Table~\ref{tab:zappos-classification} and Table~\ref{tab:medic-classification}. For the Zappos50k dataset, MSCon has the highest top-1 classification accuracy of the models. For MEDIC, MSCon out performs all of the contrastive learning techniques on all tasks. However, for three of the tasks, the best performance is achieved by one of cross-entropy methods (but different methods dominate for different tasks). We hypothesize that this may be due to the inherent uncertainty of some of the tasks \cite{alam2018crisismmd, alam2022medic}. For both datasets, CSN achieves accuracies that are lower than the single-task supervised networks. We believe this is because conditional similarity loss is trained with triplet loss \cite{hoffer2015deep}, which has been shown to be outperformed by N-pairs loss and supervised contrastive learning for single-task learning \cite{sohn2016improved, khosla2020supervised}. 

\begin{table}[!htbp]
    \centering
    \caption{\textbf{Top-1 classification accuracy across all in-domain evaluation settings for the Zappos50k dataset.} MSCon outperforms all baselines.}\label{tab:zappos-classification}
    \begin{tabular}{rccc}
      \toprule 
      \bfseries  & \multicolumn{3}{c}{\textbf{Zappos50k: In-Domain Evaluation}} \bfseries \\
      \midrule 
      \bfseries Loss & Category & Closure & Gender \bfseries \\
      XEnt Cat & 96.64 (0.34) & 74.55 (0.38) & 63.78 (0.59) \\
      XEnt Clo & 88.99 (0.33) & 92.28 (0.35) & 66.59 (0.57) \\
      XEnt Gend & 81.96 (0.32) & 73.28 (0.37) & 83.09 (0.60) \\
      XEnt MT & 96.98 (0.29) & 93.33 (0.36) & 85.07 (0.55) \\
      \midrule
      SimCLR & 90.05 (0.43) & 81.30 (0.49) & 69.10 (0.84) \\
      SupCon Cat & 96.95 (0.29) & 73.02 (0.36) & 61.24 (0.62) \\
      SupCon Clo & 83.62 (0.30) & 91.75 (0.41) & 65.90 (0.60) \\
      SupCon Gen & 76.40 (0.28) & 69.52 (0.38) & 85.11 (0.58) \\
      CSN & 83.33 (0.32) & 72.12 (0.36) & 69.21 (0.60) \\
      MSCon & \textbf{97.17 (0.27)} & \textbf{94.37 (0.35)} & \textbf{85.98 (0.56)} \\
      \bottomrule 
    \end{tabular}
\end{table}

\begin{table*}[!htbp]
    \centering
    \caption{\textbf{Top-1 classification accuracy across all in-domain evaluation settings for the MEDIC dataset.} We compare cross-entropy single-task and multi-task training, unsupervised contrastive training (SimCLR), single-similarity contrastive training (SupCon), and multi-similarity contrastive training (CSN, MSCon). }\label{tab:medic-classification}
    \begin{tabular}{rcccc}
      \toprule 
      \bfseries  & \multicolumn{4}{c}{\textbf{MEDIC: In-Domain Evaluation}} \bfseries \\
      \midrule 
      \bfseries Loss & Damage severity & Disaster types & Humanitarian & Informative \bfseries \\
      XEnt Damage severity & \textbf{81.39 (0.35)} & 75.71 (0.37) & 81.76 (0.33) & 84.48 (0.31) \\
      XEnt Disaster types & 81.02 (0.34) & 78.98 (0.35) & 82.06 (0.34) & 86.08 (0.3) \\
      XEnt Humanitarian & 81.32 (0.36) & 76.52 (0.35) & \textbf{82.1 (0.37)} & \textbf{86.41 (0.31)} \\
      XEnt Informative & 80.2 (0.36) & 76.73 (0.35) & 80.83 (0.36) & 85.68 (0.3) \\
      XEnt Multi-Task & 81.01 (0.36) & 78.04 (0.32) & \textbf{82.25 (0.35)} & 86.01 (0.29) \\
      \midrule
      SimCLR & 74.9 (0.4) & 68.5 (0.42) & 73.89 (0.4) & 78.67 (0.33) \\
      SupCon Damage severity & 80.26 (0.33) & 75.1 (0.4) & 80.42 (0.4) & 84.45 (0.34) \\
      SupCon Disaster types & 80.23 (0.34) & 78.33 (0.37) & 80.63 (0.36) & 84.02 (0.3) \\
      SupCon Humanitarian & 79.98 (0.36) & 74.89 (0.39) & 80.36 (0.32) & 85.07 (0.32) \\
      SupCon Informative & 79.14 (0.35) & 74.67 (0.34) & 79.97 (0.31) & 84.02 (0.3) \\
      CSN & 75.13 (0.4) & 70.02 (0.37) & 70.52 (0.38) & 76.28 (0.32) \\
      MSCon & 81.0 (0.3) & \textbf{79.14 (0.31)} & 81.69 (0.3) & 85.15 (0.3) \\
      \bottomrule 
    \end{tabular}
\end{table*}

\textbf{Out-of-domain Performance.} Here, we test how well different approaches are able to generalize to previously unseen tasks. We compare MSCon to multi-task cross-entropy (XEnt MT). For the Zappos50k dataset, we train embedding spaces with the category, closure, and gender similarity metrics. We then select the top 20 brands in the Zappos dataset with the most examples, and fine-tune a classification layer on the frozen embedding with the brand labels. We report top-1 brand classification accuracy of the fine-tuned network on the test set and the standard deviation in Table~\ref{tab:zappos-gen}. We find that MSCon significantly improves upon XEnt MT in the out-of-domain setting. More detailed top-1 classification results for all cross-entropy and contrastive networks are provided in the supplement.

To evaluate generalization on the MEDIC dataset, we hold out each of the four tasks. We then train an embedding space with the remaining three similarity metrics. Next, we fine-tune a classification layer on the frozen embedding with the hold-out task. Table~\ref{tab:medic-gen} reports the top-1 classification accuracy and the standard deviation for the hold-out task on the test set.  We observe that, except for the informative task, our approach is able to generalize to new tasks with higher accuracy than the multi-task cross-entropy learned embedding space. We hypothesize that this is because the informative task is the only binary task and the most ambiguous.

\begin{table*}
    \parbox{0.3\linewidth}{
    \centering
    \caption{\textbf{Top-1 out-of-domain brand classification accuracy for the Zappos50k dataset.} We compare cross-entropy multi-task training and multi-similarity contrastive training. }\label{tab:zappos-gen}
    \begin{tabular}{rc}
      \toprule 
      \multicolumn{2}{c}{\textbf{Zappos50k: OOD Evaluation}} \\
      \midrule 
      \bfseries Loss & Brand \bfseries \\
      XEnt MT & 32.10 (1.48) \\
      MSCon & \textbf{42.62 (1.52)} \\
      \bottomrule 
    \end{tabular}}
\hfill
\parbox{0.65\linewidth}{
    \centering
    \caption{\textbf{Top-1 out-of-domain classification accuracy for hold-out similarity metrics on the MEDIC dataset.} We compare cross-entropy multi-task training and multi-similarity contrastive training. We abbreviate the similarity metrics of damage severity (DS), disaster types (DS), humanitarian (Human), informative (Inf) and multi-task (MT) and MSCon.}\label{tab:medic-gen}
    \begin{tabular}{rcccc}
      \toprule 
      \bfseries  & \multicolumn{4}{c}{\textbf{MEDIC: OOD Evaluation}} \bfseries \\
      \midrule 
      \bfseries Loss & DS & DT & Human & Inf \bfseries \\
      XEnt MT & 79.51 (0.36) & 75.02 (0.38) & 79.77 (0.4) & \textbf{86.18 (0.3)} \\
      MSCon & \textbf{80.98 (0.32)} & \textbf{76.17 (0.32)} & \textbf{81.45 (0.34)} & 85.22 (0.3) \\
      \bottomrule 
    \end{tabular}}
\end{table*}
    
\begin{figure*}[!htbp]
\centering
\centerline{\includegraphics[width=2\columnwidth]{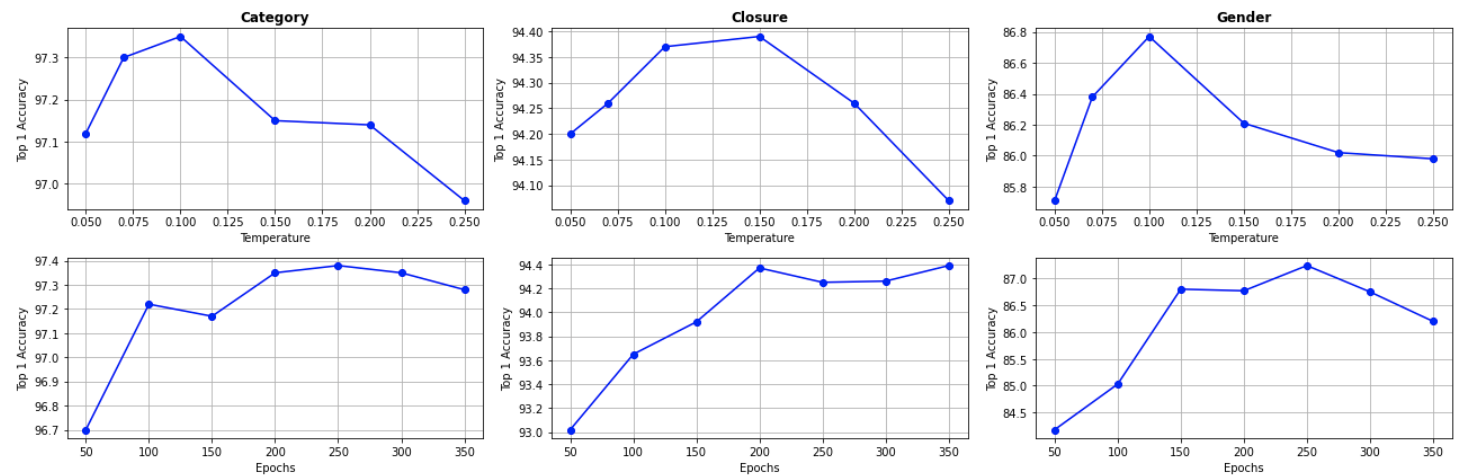}}
\caption{\textbf{Hyperparameter effect on top-1 accuracy for the Zappos50k dataset.} The top row shows top-1 classification accuracy as a function of temperature during pretraining stage for MSCon. The bottom row shows top-1 classification accuracy as a function of pretraining epochs for MSCon.}
\label{fig:zappos-hyper}
\end{figure*}

\paragraph{Hyperparameter Analysis.} We test if there exists a specific temperature that leads to optimal performance of MSCon for multiple similarity metrics. In Figure~\ref{fig:zappos-hyper}, we plot the top-1 classification accuracy for each of the category, closure, and gender tasks as a function of pretraining temperature for MSCon. We also plot the top-1 classification accuracy as a function of training epochs. We find that a pretraining temperature of $\tau=0.1$ and training for 200 epochs works well for all tasks. These hyperparameter settings are consistent with optimal hyperparameter settings for SimCLR and SupCon. Note that previous work for SimCLR and SupCon have found the large batch sizes consistently result in better top-1 accuracy \cite{chen2020simple, khosla2020supervised}. We hypothesize that larger batch sizes would also improve performance for MSCon loss. We include hyperparameter analyses on MSCon for the MEDIC dataset in the supplement.

\section{Conclusion}\label{sec:conc}

In this work, we propose multi-similarity contrastive loss (MSCon). Existing contrastive learning methods learn a representation based on a single similarity metric. However, it is often the case that multiple tasks are available, each implying a different similarity metric.  We show how to leverage multiple similarity metrics in a contrastive setting to learn embeddings that generalize well to unseen tasks. We additionally extend uncertainty based task weighting to the contrastive framework. We do this by 1) modeling downstream classification performance for each similarity by using a psuedo-likelihood and 2) by representing similarity dependent uncertainty as a temperature scaling factor for each similarity metric. We demonstrate that our MSCon learned embeddings outperform all contrastive baselines and generalizes better than multi-task cross-entropy to novel tasks.

There are many interesting directions for future work. Firstly, we do not consider data-dependent uncertainty in our framework. It would be interesting to consider what would happen if we have variance in the uncertainty of our input data. Can we account for both similarity-dependent and input-dependent uncertainty? Another interesting direction for future work would be to see if we could incorporate non-categorical labels in our multi-similarity learning scheme. Currently, we define similarity metrics using multiple categorical tasks. However, some applications use continuous metrics of similarity (e.g. heart rate measurements available in patient electronic health record data or heel height associated with shoes). Defining positive and negative examples for continuous variables with different scales is not straightforward. Thus, an interesting follow-up question may be how to incorporate both categorical and continuous similarity metrics under a single contrastive framework.

Finally, we note that our method will not necessarily generalize well to any novel task. Sometimes, multi-task learning can degrade performance when models are unable to learn representations that generalize towards all tasks \cite{fifty2021efficiently, zhang2021survey}. Our work does not address criteria for the selection of tasks for training or evaluation.


{\small
\bibliographystyle{ieee_fullname}
\bibliography{cvpr_2023}
}

\end{document}